\documentclass{article}
\usepackage{ijcai17}
\usepackage[ruled]{algorithm2e}
\usepackage{amsfonts}
\usepackage{amssymb,amsmath,cool}
\usepackage{graphicx}
\usepackage{subfigure}
\usepackage{stfloats}
\usepackage{ntheorem}
\usepackage{color}
\usepackage{url}
\usepackage{multirow}
\usepackage[small]{caption}
\usepackage{times}
\DeclareMathOperator*{\argmax}{argmax}
\DeclareMathOperator*{\argmin}{argmin}

\usepackage{times}
\title{Boosted Zero-Shot Learning with Semantic Correlation Regularization}
\author{Te~Pi$^1$,~~Xi~Li$^{1,2}$\footnotemark[1],~~Zhongfei~(Mark)~Zhang$^1$\\
${}^1$Zhejiang University, Hangzhou, China\\
${}^2$Alibaba-Zhejiang University Joint Institute of Frontier Technologies, Hangzhou, China\\
peterpite@zju.edu.cn;~~xilizju@zju.edu.cn;~~zhongfei@zju.edu.cn
}

\begin{document}

\maketitle

{\renewcommand{\thefootnote}{\fnsymbol{footnote}}
\footnotetext[1]{Corresponding author}}

\begin{abstract}
We study zero-shot learning (ZSL) as a transfer learning problem, and focus on the two key aspects of ZSL, model effectiveness and model adaptation. For effective modeling, we adopt the boosting strategy to learn a zero-shot classifier from weak models to a strong model. For adaptable knowledge transfer, we devise a Semantic Correlation Regularization (SCR) approach to regularize the boosted model to be consistent with the inter-class semantic correlations. With SCR embedded in the boosting objective, and with a self-controlled sample selection for learning robustness, we propose a unified framework, Boosted Zero-shot classification with Semantic Correlation Regularization (BZ-SCR). By balancing the SCR-regularized boosted model selection and the self-controlled sample selection, BZ-SCR is capable of capturing both discriminative and adaptable feature-to-class semantic alignments, while ensuring the reliability and adaptability of the learned samples. The experiments on two ZSL datasets show the superiority of BZ-SCR over the state-of-the-arts.



\end{abstract}

\section{Introduction}
Recent years have witnessed a widespread increase of interest in Zero-Shot Learning (ZSL), which aims at
learning a classifier from the data of the seen classes $\mathcal{Y}_S$ for classifying the data from the unseen target classes $\mathcal{Y}_T$.
A common solution of ZSL is to introduce some auxiliary information on labels (e.g., attributes~\cite{Akata2013SJE}, word vector representations~\cite{Frome2013DeViSE}) to model the semantic relationships between $\mathcal{Y}_S$ and $\mathcal{Y}_T$ in a common structured embedding space.
Hence, ZSL is typically solved as a transfer learning problem, which seeks to exploit the shared knowledge among
classes and transfer these knowledge to adapt on $\mathcal{Y}_T$.

From the perspective of transfer learning, a zero-shot classifier is expected to be both discriminative for the data distribution of $\mathcal{Y}_S$ and with robust adaptation to the data distribution of $\mathcal{Y}_T$. Therefore, the classifier should not only capture the significant feature-to-class semantic alignments, but also reflect the semantic connections between $\mathcal{Y}_S$ and $\mathcal{Y}_T$. In brief, the two key principles of learning a zero-shot classifier are the model effectiveness and the model adaptation.

First, for effective modeling, we adopt the boosting strategy~\cite{zhou2012ensemble} as the basis of the ZSL framework, by learning a zero-shot boosting classifier from weak models to a strong model. The superiority of boosting lies in its flexible piecewise approximation to the data distributions based on ensembles of weak hypotheses. Thus, a boosting ZSL model learns in an asymptotic way to sufficiently capture the discriminative patterns in the feature space and their alignments to the semantic patterns in the embedding space, and obtains the learning effectiveness. On the other hand, for a better adaptation to the target classes $\mathcal{Y}_T$, the model needs to explore the shared semantic patterns of $\mathcal{Y}_S$ and $\mathcal{Y}_T$ to reflect their semantic correlations. For this sake, it is reasonable to assume that a well-adapted classifier should generate a high compatibility score for a sample and a target class semantically close to the sample's true class, and a low score otherwise. Based on these analysis, we propose a Semantic Correlation Regularization (SCR) approach to impose this constraint, which encourages a negative correlation between a sample's scores on the target classes and the classes' semantic divergences to the sample's ground truth. In this way, SCR constrains the learned model to be consistent with the source-target semantic correlations, for a semantically adaptable model transfer.

Furthermore, in addition to the label relations, a feasible model adaptation also relies on a modeling of the sample relations for an exploration of the common geometric structures of the feature space and the semantic space. This requires a control or selection scheme for the samples to be learned, to distinguish the samples with robust patterns for classifying $\mathcal{Y}_S$, and with adaptable patterns for the transfer to the semantics of $\mathcal{Y}_T$. Therefore, a self-controlled sample selection of~\cite{zhao2015self} is applicable for ZSL, which adaptively incorporates the samples into learning from easy ones to complex ones, inspired by the human learning process. Such sample selection smoothly controls the learning pace of ZSL by what the model has already learned. With this self-controlled learning pace, a zero-shot model is smoothly guided to focus on those both reliable and adaptable samples to explore the robust feature-semantic alignments, for learning a classifier with good generalization to the target classes.

Therefore, motivated by a simultaneous enhancement of the model effectiveness and model adaptation for ZSL as a transfer learning task, we propose a novel framework, Boosted Zero-shot classification with Semantic Correlation Regularization (BZ-SCR). With the SCR embedded into the boosted classification, the boosting process would favor models with not only accurate prediction for an effective classifier, but also with semantic consistency to the target classes for an adaptable model transfer. With a self-controlled sample selection, the BZ-SCR puts emphasis on the reliable and adaptable samples to explore the common feature-semantic structures for a robust model generalization. The proposed framework learns by effectively balancing boosted model selection and self-controlled sample selection. As a result, the proposed framework is capable of jointly considering the learning effectiveness, the cross-semantics adaptation, and the model robustness for learning a zero-shot classifier.

In mathematics, we formulate BZ-SCR as a max-margin boosting optimization with self-controlled sample selection. The contributions of this paper are summarized as follows:

1. We propose a boosted ZSL approach that theoretically seeks for effective knowledge transfer in both model learning effectiveness and cross-domain model adaptation, via max-margin boosting optimization with self-controlled sample selection. To the best of our knowledge, this work is innovative in ZSL by jointly considering model effectiveness, model adaptation, and sample selection within a boost learning framework.

2. We present a novel Semantic Correlation Regularization (SCR) for ZSL, which regularizes the learned boosting model to be consistent with the source-target semantic correlations. In principle, the proposed SCR is capable of effectively capturing the inter-class correlations by modeling the intrinsic geometrical structure properties.

\section{Related Work}
We review the related work on zero-shot learning, boost learning and self-controlled sample selection approaches.

Known as learning from disjoint training and test classes, ZSL requires the ability to transfer knowledge from classes with training data to classes without. The possible sources of side informations of classes include manually annotated attributes~\cite{Akata2013SJE}, semantic class taxonomies~\cite{wordnet1995}, and unsupervised word representations~\cite{Mikolov2013word2vec}. Based on the way of leveraging these informations, the existing ZSL approaches fall into two categories. One is to build intermediate attribute classifiers and use their probabilistic weightings to make class predictions~\cite{Lampert2014tpami,Lampert2009cvpr}. The other group is to represent the semantic knowledge of labels in an embedding space and directly learn a mapping function between the feature space and the embedding space. Among them, CCA~\cite{hastie2001elements} maximizes the inter-domain statistical correlations; \cite{Palatucci2009nips} learns a linear mapping, ALE~\cite{Akata2016tpamiALE}, SJE~\cite{Akata2015cvprSJE} and DeViSE~\cite{Frome2013DeViSE} learn a bilinear mapping, and~\cite{Socher2013nips} models a nonlinear regression with neural networks. Efforts for directly tackling the absence of the training data are also made, such as semi-supervised transductive methods~\cite{Guo2016Transductive} and the generation of labeled virtual data for unseen classes~\cite{Wang2016aaaiRKT}.

Boosting is a family of supervised ensemble approaches that build a strong model by successively learning and combining multiple weak models~\cite{zhou2012ensemble}. The main variation among boosting methods is their ways of weighting samples and weak learners, e.g., Adaboost~\cite{freund1997decision}, SoftBoost~\cite{ratsch2007boosting} and LPBoost~\cite{LPBoost}. The superiority of boosting lies in its piecewise approximation of a nonlinear decision function to explore the data patterns~\cite{schapire2012boosting}.

Proposed by~\cite{kumar2010self}, the self-controlled sample selection is inspired by the learning process of humans that gradually incorporates the training samples into learning from easy ones to complex ones. It is initially developed for avoiding the bad local minima of latent models, by learning the data in an order from easy to hard determined by the feedback of the learner itself. It is applied in different applications, such as multimedia reranking~\cite{jiang2014easy}, matrix factorization~\cite{zhao2015self}, and multiple instance learning~\cite{zhang2015MIP}. \cite{meng2015objective} provides a theoretical analysis of the robustness of this scheme, which reveals its consistency with the non-convex upper-bounded regularization.


\section{Our Approach} \label{Sec_MF}

\subsection{Problem Formulation} \label{SSec_SPBF}

Let $\left\{\left(x_i,y_i\right)\right\}_{i=1}^N$ be a set of $N$ training samples with feature $x_i\in\mathcal{X}\subseteq\mathbb{R}^m$ and label $y_i\in\mathcal{Y}_S=\left\{1,\ldots,C_S\right\}$, where $\mathcal{Y}_S$ denotes the set of $C_S$ seen labels. The goal of ZSL is to learn a classifier for a target label set $\mathcal{Y}_T=\left\{C_S+1,\ldots,C\right\}$ disjoint from $\mathcal{Y}_S$. Let $\mathcal{Y}=\mathcal{Y}_S\cup\mathcal{Y}_T$. The side information of labels such as attributes is available to embed the labels into a structured embedding space, represented by an embedding function $\varphi:\mathcal{Y}\rightarrow\mathbb{R}^d$. As common in supervised learning, the classifier aims to learn a compatibility score function $F:\mathcal{X}\times\mathcal{Y}\rightarrow\mathbb{R}$ with which the prediction is made:
\begin{equation}
\tilde{y}\left(x\right)=\argmax_{r\in\mathcal{Y}}{F\left(x,r;w\right)},
\end{equation}
where $F\left(x,r;w\right)$ determines the compatibility of the pair $(x,r)$ based on the label embedding $\varphi\left(r\right)$ and parameter $w$. The general max-margin formulation for zero-shot classification with loss $L$ and regularization $\Omega$ is given by:
\begin{gather} \label{gen_class}
\min_{w}{\sum_{i=1}^N{\sum_{r\in\mathcal{Y}_S}{L\left(\rho_{ir}\right)}}+\nu\Omega\left(w\right)}\\
s.t.~\forall i,r, \rho_{ir}={\delta F}\left(x_i,r,y_i;w\right)+\Delta(y_i,r),\notag
\end{gather}
where ${\delta F}\left(x_i,r,y_i;w\right)=F\left(x_i,r;w\right)-F\left(x_i,y_i;w\right)$; $\rho_{ir}$ is the score margin of $x_i$ between class $r$ and its ground truth $y_i$; $\Delta:\mathcal{Y}\times\mathcal{Y}\rightarrow[0,1]$ defines the semantic divergence between two labels, as a penalty of predicting $r$ for the true label $y_i$; $\nu>0$ is a trade-off hyperparameter. Generally, the loss function $L:\mathbb{R}\rightarrow\mathbb{R}^+$ should be convex and monotonically increasing for a small $\rho$.

\subsection{Semantic Correlation Regularization} \label{SSec_SCR}
In Section~\ref{SSec_SPBF}, we have considered to learn a max-margin classifier on $\mathcal{Y}_S$. However, due to the divergences of semantics and data distributions between $\mathcal{Y}_S$ and $\mathcal{Y}_T$, the learned models from $\mathcal{Y}_S$ may not necessarily adapt well on $\mathcal{Y}_T$. Thus, we aim to regularize the model based on the source-target semantic correlations for better cross-semantics adaptation.

First, we notice that the semantic correlations of a sample $(x_i,y_i)$ to the classes of $\mathcal{Y}_T$ are not uniform, but specified by $\Delta(y_i,r)$. Hence, it is reasonable to assume that a sample $(x_i,y_i)$ should be assigned a high score $F(x_i,r)$ for class $r\in\mathcal{Y}_T$ if $r$ is semantically close to $y_i$ ($\Delta(y_i,r)$ is low), and a low score $F(x_i,r)$ if $r$ is semantically far from $y_i$ ($\Delta(y_i,r)$ is high). In other words, for a pair of classes $r_1,r_2\in\mathcal{Y}_T$, $F(x_i,r_1)$ is expected larger than $F(x_i,r_2)$ if $\Delta(y_i,r_1)$ is smaller than $\Delta(y_i,r_2)$, and vice versa. Thus, we consider the following term:
\begin{equation}
\sigma_i(r_1,r_2)\triangleq\left[\Delta(y_i,r_1)-\Delta(y_i,r_2)\right]\left[F(x_i,r_1)-F(x_i,r_2)\right],   \notag
\end{equation}
and penalizes $\max{(0,\sigma_i(r_1,r_2))}$ that is valid if $\sigma_i(r_1,r_2)>0$, i.e., the magnitude relationship of the two scores are conflicted with that of their $\Delta$ divergences. Then, by the summation over $r_1,r_2\in\mathcal{Y}_T$, and with a relaxation of a sum of hinge to a hinge of sum for ease of optimization, we have:
\begin{align}
&\sum\nolimits_{r_1,r_2\in\mathcal{Y}_T}{\max{(0,\sigma_i(r_1,r_2))}}\notag\\
\xrightarrow{\text{relax}}&\max{\left(0,\sum\nolimits_{r_1,r_2\in\mathcal{Y}_T}{\sigma_i(r_1,r_2)}\right)}   \notag\\
=&\max{(0,2{|\mathcal{Y}_T|}^2\left\{\text{E}[\Delta_i^t\odot F_i^t]-\text{E}[\Delta_i^t]\text{E}[F_i^t]\right\})}\notag\\
=&2{|\mathcal{Y}_T|}^2\max{\left(0,\text{cov}\left[\Delta_i^t,F_i^t\right]\right)}\triangleq2{|\mathcal{Y}_T|}^2\max{(0,{cov}_i)}, \notag
\end{align}
where $\Delta_i^t,F_i^t\in\mathbb{R}^{|\mathcal{Y}_T|}$ are the stacked vectors of $\Delta(y_i,r)$ and $F(x_i,r)$ for $r\in\mathcal{Y}_T$, respectively; $\text{E}[\cdot]$ is a mean operator and $\text{cov}[\cdot,\cdot]$ is a covariance operator on the elements of vectors.

Based on the above analysis, we propose the Semantic Correlation Regularization (SCR), to encourage the covariance ${cov}_i$ to be lower than $0$ such that the score distributions and the semantic divergence distributions on $\mathcal{Y}_T$ are negatively correlated. Instead of using $\max{(0,{cov}_i)}$, we use a smooth surrogate of the hinge function, the logistic function, for derivation convenience. The SCR is defined as:
\begin{equation} \label{form_SCR}
R(\rho_i^t)\triangleq\ln{(1+e^{{cov}_i})}, ~R:\mathbb{R}^{|\mathcal{Y}_T|}\rightarrow\mathbb{R},
\end{equation}
where $\rho_i^t\in\mathbb{R}^{|\mathcal{Y}_T|}$ is the stacked vector of $\rho_{ir}$ for $r\in\mathcal{Y}_T$. Here, we define the SCR term as a function of $\rho_i^t$ for the convenience of dual derivation in Section~\ref{SSec_Opt}, since
\begin{align}
{cov}_i&=\text{cov}\left[\Delta_i^t,F_i^t\right]=\text{cov}\left[\Delta_i^t,F_i^t-F(x_i,y_i)\right] \notag\\
&=\text{cov}\left[\Delta_i^t,\rho_i^t-\Delta_i^t\right]=\text{cov}\left[\Delta_i^t,\rho_i^t\right]-\text{D}[\Delta_i^t]. \notag
\end{align}

With the SCR embedded into the classification objective Eq.~(\ref{gen_class}), the learned model is regularized to be consistent with the source-target semantic correlations, which improves the model adaptation to the target classes.

\subsection{BZ-SCR Framework}

We formulate the BZ-SCR framework based on the boost learning, the SCR in Section~\ref{SSec_SCR}, and the self-controlled sample selection, for an effective and adaptable zero-shot classifier. Specifically, for effective modeling, we adopt the boosting strategy to formulate $F$ as an ensemble of weak classifiers $\left\{h_j\in\mathcal{H}\right\}_{j=1}^K$ in the space of weak models $\mathcal{H}$:
\begin{equation} \label{boost_F}
F\left(x,r;w\right)=\sum_{j=1}^K{w_j h_j\left(x,r\right)},~w\geqslant0,
\end{equation}
where each $h_j:\mathcal{X}\times\mathcal{Y}\rightarrow\mathbb{R}$ is a base score function; $w$ is specified as the weight parameter to be learned.

On the other hand, the samples relations should also be modeled in addition to the labels relations for a robust model transfer. Thus, inspired by the adaptive scheme of~\cite{kumar2010self} that learns a model smoothly from the easy/faithful samples to the hard/confusing ones, we reformulate the objective of Eq.~(\ref{gen_class}) with a self-controlled sample selection procedure. Then, based on the boosted model Eq.~(\ref{boost_F}) and the SCR Eq.~(\ref{form_SCR}), we have the formulation of BZ-SCR framework:
\begin{gather} \label{final_SPB}
\min_{w,s}{\sum_{i=1}^N{\left\{s_i\left[\sum_{r\in\mathcal{Y}_S}{L\left(\rho_{ir}\right)}+\beta R(\rho_i^t)\right]+g\left(s_i;\lambda\right)\right\}}+\nu\Omega(w)},\\
s.t.~\forall i,r,\rho_{ir}={\delta F}\left(x_i,r,y_i\right)+\Delta(y_i,r);w\geqslant0;s\in{[0,1]}^N,\notag
\end{gather}
where $s_i\in[0,1]$ is the weight of sample $x_i$ that indicates its learning ``easiness"; $g\left(\cdot;\lambda\right):\left[0,1\right]\rightarrow\mathbb{R}$ is the function that specifies how the samples are selected (the reweighting scheme of $s$) controlled by the parameter $\lambda>0$; and $\beta>0$ is a trade-off hyperparameter between the classification loss $L(\cdot)$ for fitness on $\mathcal{Y}_S$ and the SCR $R(\cdot)$ for adaptation to $\mathcal{Y}_T$.
Note that in Eq.~(\ref{final_SPB}), a weight $s_i$ is assigned to each sample as a measure of its ``easiness", which is tuned based on the currently learned confidence (including classification loss and SCR) and the function $g\left(s_i;\lambda\right)$ to adaptively select reliable and adaptable samples.

For the convenience of derivation, we specify the loss $L(\cdot)$ as a smooth loss function, the logistic loss, and specify the regularization $\Omega(\cdot)$ as the $l_1$-norm to impose a sparsity constraint on the model ensemble:
\begin{equation} \label{L_Reg}
L\left(\rho\right)\triangleq\ln{\left(1+e^\rho\right)};~\Omega\left(w\right)\triangleq{\left\|w\right\|}_1=\sum\nolimits_j{w_j}.
\end{equation}

\subsection{Optimization} \label{SSec_Opt}
Following~\cite{pi2016SPBL}, we use the alternating optimization to solve Eq.~(\ref{final_SPB}). For the optimization of $s$, we have
\begin{equation} \label{opt_s}
s_i^*=\argmin_{s_i}{s_i l_i+g\left(s_i;\lambda\right)},~s.t.~s_i\in{\left[0,1\right]},
\end{equation}
where $l_i=\sum_{r\in\mathcal{Y}_S}{L(\rho_{ir})}+\beta R(\rho_i^t)$ is a composite cost of model fitness on $\mathcal{Y}_S$ and model transfer utility on $\mathcal{Y}_T$. Based on the summarized general properties and candidates of the $g$ function in~\cite{jiang2014easy}, we specify $g$ and the corresponding $s_i^*$ as the scheme for mixture weighting due to its better overall performance in the experiments:
\begin{gather}
g\left(s_i;\lambda,\zeta\right)=-\zeta\ln{\left(s_i+{\zeta}/{\lambda}\right)},~\lambda,\zeta>0,\notag\\
s_i^*=\left\{\begin{array}{rcl} 1,~~~~~~~~~&l_i\leqslant {\zeta\lambda}/{\left(\zeta+\lambda\right)} \label{update_s}\\
0,~~~~~~~~~&l_i\geqslant\lambda \\
{\zeta}/{l_i}-{\zeta}/{\lambda},~&\mbox{otherwise}
\end{array}\right. ,
\end{gather}
which is a mixture of hard 0-1 weighting and soft real-valued weighting, with an extra parameter $\zeta$. Note that $s_i^*$ is monotonically decreasing with $l_i$ and increasing with $\lambda$, so as to select easy samples with small losses under the tolerance $\lambda$.

For the optimization of $w$, we have
\begin{gather} \label{opt_w}
w^*=\argmin_{w}{\sum_{i}{s_il_i(\rho_i)}+\nu{\left\|w\right\|}_1},\\
s.t.~\forall i,r,\rho_{ir}={\delta F}\left(x_i,r,y_i\right)+\Delta(y_i,r);~w\geqslant0.\notag
\end{gather}

To solve $w$ in Eq.~(\ref{opt_w}), we adopt the column generation method~\cite{LPBoost} in the dual space of $w$ to handle the potentially infinite candidate weak models in the $\mathcal{H}$ space. We check the dual problem of Eq.~(\ref{opt_w}):
\begin{gather} \label{dual_opt}
\min_Q{\sum_{i,r\in\mathcal{Y}_S}{H(Q_{ir})}+\sum_i{\beta R_{s_i}^\star\left({Q_i^t}/{\beta}\right)}-J(Q,\Delta)},\\
s.t.~\sum_{i,r}{Q_{ir}\left(h^{(iy_i)}-h^{(ir)}\right)}\leqslant \nu\textbf{1}_K,\notag
\end{gather}
where $Q\in\mathbb{R}^{N\times C}$ is the Lagrangian multiplier of the equality constraints of Eq.~(\ref{opt_w}); $Q_i^t\in\mathbb{R}^{|\mathcal{Y}_T|}$ is the stacked vector of $Q_{ir}$ for $r\in\mathcal{Y}_T$; $R_{s_i}^\star:\mathbb{R}^{|\mathcal{Y}_T|}\rightarrow\mathbb{R}$ is the Fenchel dual function of $s_iR(\cdot)$; $H(Q_{ir})=Q_{ir}\ln{Q_{ir}}+(s_i-Q_{ir})\ln{(s_i-Q_{ir})}$; $J(Q,\Delta)=\sum_{i,r}{Q_{ir}\Delta(y_i,r)}$; $h^{(ir)}\in\mathbb{R}^K$ is the stacked vector of $h_j(x_i,r)$ for $j\in[1,K]$. The relation between the dual and the primal solution is:
\begin{equation} \label{Q_rho}
Q_{ir}=\left\{\begin{array}{rcl} \frac{s_i}{1+\exp{(-\rho_{ir})}},~~~~~~~~~~~~&r\in\mathcal{Y}_S\\
\frac{\Delta(y_i,r)-\text{E}[\Delta_i^t]}{|\mathcal{Y}_T|}\frac{\beta s_i}{1+\exp{(-{cov}_i)}},&r\in\mathcal{Y}_T
\end{array}\right. .
\end{equation}
Please refer to Appendix~\ref{apdx_dual} for the dual derivation of Eqs.~(\ref{dual_opt}) and~(\ref{Q_rho}).

Based on the column generation, the set of weak models is augmented by a weak model $\hat{h}$ that most violates the current dual constraint in Eq.~(\ref{dual_opt}):
\begin{equation} \label{new_h}
\hat{h}=\argmax_{h\in\mathcal{H}}{\sum_{i,r}{Q_{ir}\left\{h(x_i,y_i)-h(x_i,r)\right\}}}.
\end{equation}
Then the optimization continues with the new set of weak models, until the violation score (objective value of Eq.~(\ref{new_h})) reaches a tolerance threshold.

Eq.~(\ref{new_h}) indicates that the matrix $Q$ serves as the sample importance for learning a new weak model. From Eq.~(\ref{Q_rho}), we see that $Q$ gives high weights to not only the misclassified samples on $\mathcal{Y}_S$ (with large $\rho_{ir}$) and those not aligned well with their semantic correlations on $\mathcal{Y}_T$ (with large positive ${cov}_i$), but also the reliable samples with high weights $s_i$. These reliable samples are those both discriminative on $\mathcal{Y}_S$ and adapt well on $\mathcal{Y}_T$ in the previous iteration. As a result, the future weak learners will emphasize on samples both insufficiently learned currently and easily learned previously, so as to obtain an effective, adaptable and robust zero-shot classifier for knowledge transfer to the target classes.


We summarize the optimization procedure in Algorithm~\ref{algo_BZ_SCR}. Note that the sample selection parameters $\left(\lambda,\zeta\right)$ are iteratively increased (annealed) when they are small (Line~\ref{algo_anneal_1} to~\ref{algo_anneal_2}), so as to introduce more complex samples in the future learning. An early stopping criterion is adopted to maintain a reasonable running time.


\begin{algorithm}[tp]
\caption{Boosted Zero-shot classification with Semantic Correlation Regularization (BZ-SCR)\label{algo_BZ_SCR}}
\LinesNumbered
\SetKwInOut{Input}{Input} \SetKwInOut{Output}{Output} \SetKwInOut{Return}{Return}
\SetKwInOut{Validation}{Validation} \SetKwInOut{Annealing}{Annealing}
\Input{Training set ${\left\{\left(x_i,y_i\right)\right\}}_{i=1}^N$; label embeddings $\{\varphi(y)\}_{y\in\mathcal{Y}}$; $\nu$; $\beta$; $\lambda_{max}$; $T_{ES}$; $\mu>1$; $\epsilon>0$.}
\Output{A set of $K$ weak models ${\left\{h_j\in\mathcal{H}\right\}}_{j=1}^K$; $w$.}
\textbf{Initialize}: $s^{(0)}$; $(\lambda,\zeta)$; $Q\leftarrow s^{(0)}\mathbf{1}_C^T$; $t\leftarrow 0$\;
\Repeat{
$\sum\nolimits_{i,r}{Q_{ir}\left\{h_t(x_i,y_i)-h_t(x_i,r)\right\}}<\nu+\epsilon$
\textbf{or}
$t\geqslant T_{ES}$ \textbf{and} $err^{\left(t\right)}>\min\nolimits_{1\leqslant\tau\leqslant t-1}{err^{\left(\tau\right)}}$
}{
$t\leftarrow t+1$\;
Learn a new weak model: solve Eq.~(\ref{new_h}) to obtain $h_t$ based on $Q$\;\label{algo_newh}
Update $w$: solve Eq.~(\ref{opt_w}) for $w^{\left(t\right)}$ based on $s^{\left(t-1\right)}$\;\label{algo_w}
Update $Q$: compute $Q$ by Eq.~(\ref{Q_rho}) based on $w^{\left(t\right)}$ and $s^{\left(t-1\right)}$\;\label{algo_Q}
Update $s$: compute $s^{\left(t\right)}$ by Eq.~(\ref{update_s}) based on $w^{\left(t\right)}$\;\label{algo_s}
\Validation{}
Test ${\left\{h_j\right\}}_{j=1}^t$ and $w^{\left(t\right)}$ on the validation set, to obtain the classification error rate $err^{\left(t\right)}$\;
\Annealing{}
\If{$\lambda<\lambda_{max}$\label{algo_anneal_1}}{
$(\lambda,\zeta)\leftarrow(\mu\lambda,\mu\zeta)$\;}\label{algo_anneal_2}
}
$K\leftarrow\argmin\nolimits_\tau{err^{\left(\tau\right)}}$\;
\Return{${\left\{h_j\right\}}_{j=1}^K$, $w=w^{\left(K\right)}$.}
\end{algorithm}

Furthermore, the convergence of Algorithm~\ref{algo_BZ_SCR} is guaranteed by the convexity of the two subproblems Eqs.~(\ref{opt_s}) and~(\ref{opt_w}) of the alternating optimization. This leads to the monotonic decreasing of the objective value of Eq.~(\ref{final_SPB}) after each iteration. Since the objective is bounded below, such optimization procedure is guaranteed to converge. Figure~\ref{fig_iter}(a)(b) in the experiment empirically shows that the convergence is usually reached within several hundreds of iterations.


\section{Experiments}
We evaluate the performance of BZ-SCR on the two classic ZSL image datasets, \textit{Animal With Attributes}\footnote{http://attributes.kyb.tuebingen.mpg.de/} (\textit{AWA}) and \textit{Caltech-UCSD-Birds-200}\footnote{http://www.vision.caltech.edu/visipedia/CUB-200-2011.html} (\textit{CUB200}). The comparative methods include Structured Joint Embedding (SJE)~\cite{Akata2015cvprSJE}, Attribute Label Embedding (ALE)~\cite{Akata2016tpamiALE}, Nonlinear Regression (NR)~\cite{Socher2013nips}, and Canonical Correlation Analysis (CCA)~\cite{hastie2001elements}.

\subsection{Datasets and Experimental Settings}

\begin{table}[tp]
\center
\caption{Statistics of the datasets} \label{tab_data}
\vspace{0.05cm}
{\small
\begin{tabular}{|c|c|c|}
\hline
Dataset    &   \textit{AWA}  &  \textit{CUB200}   \\\hline
Samples    &   30475         &  11788      \\\hline
Classes (Split) & 50 (40/10) &  200 (150/50) \\\hline
Feature (dim $m$) & PCAed VGG (512) & GoogleNet (1024) \\\hline
Embedding (dim $d$) & Word2Vec (300) & Attributes (312) \\\hline
\end{tabular}
}
\end{table}

We summarize the statistics of the datasets in Table~\ref{tab_data}. For label embeddings, we use the 300-D GoogleNews Word2Vec\footnote{http://code.google.com/archive/p/word2vec/} representations for \textit{AWA}, and the 312-D annotated attributes for \textit{CUB200}. For the class split, we adopt the default split (40/10 for train+val/test) for \textit{AWA} and the same split as~\cite{Akata2013SJE} (150/50 for train+val/test) for \textit{CUB200}.

We specify for \textit{AWA} the semantic divergence $\Delta$ based on the wordnet hierarchy\footnote{http://stevenloria.com/tutorial-wordnet-textblob/}:
\begin{equation} \label{DeltaAWA}
\Delta(y_1,y_2) = 1-{(SPath(y_1,y_2)+1)}^{-1},
\end{equation}
where $SPath(y_1,y_2)$ is the length of the shortest path between words $y_1$, $ y_2$ in the wordnet hierarchy. For \textit{CUB200}, we specify $\Delta$ based on the cosine similarities between the label embeddings (attributes):
\begin{equation} \label{DeltaCUB}
\Delta(y_1,y_2) = \frac{1-cos\left(\varphi(y_1),\varphi(y_2)\right)}{1-\min\nolimits_{r_1,r_2\in\mathcal{Y}}{cos\left(\varphi(r_1),\varphi(r_2)\right)}},
\end{equation}
where $cos(\cdot,\cdot)$ is the cosine similarity between two vectors, and the denominator is for a normalization to $[0,1]$.

We specify the weak score function $h(x,y)$ as a rank-one bilinear model, for the convenience of learning a new $\hat{h}$:
\begin{equation} \label{h_form}
h(x,y;u,v) = x^Tu\cdot v^T\varphi(y),~s.t.~{\|u\|}_2={\|v\|}_2=1.
\end{equation}
With $h$ taking the form of Eq.~(\ref{h_form}), the learning of $\hat{h}$ (Eq.~(\ref{new_h})) is equivalent to a simple SVD problem.

We adopt the strategy in~\cite{jiang2014SPLD} for the annealing of the parameters $(\lambda,\zeta)$ (Line~\ref{algo_anneal_1} to~\ref{algo_anneal_2} in Algorithm~\ref{algo_BZ_SCR}). At each iteration, we sort the samples in the ascending order of their losses, and set $(\lambda,\zeta)$ based on the proportion of samples to be selected by now. We anneal the proportion of the selected samples instead of the absolute values of $(\lambda,\zeta)$, which is shown more stable in~\cite{jiang2014SPLD}.

We implement a grid search for the tuning of the hyperparameters $\nu$, $\beta$, and report the best performances.

\subsection{Experimental Results}
\begin{table}[tp]
\center
\caption{The $\bar{\Delta}$ and error rate performance of each method\label{tab_errrate}}
\vspace{0.05cm}
\begin{tabular}{|c||c|c||c|c|}
\hline
\multirow{2}*{} & \multicolumn{2}{|c||}{\textit{AWA}} & \multicolumn{2}{|c|}{\textit{CUB200}} \\\cline{2-5}
  &  \raisebox{-0.4ex}[0pt][0pt]{$\bar{\Delta}$}  &  \raisebox{-0.3ex}[0pt][0pt]{ER}  &  \raisebox{-0.4ex}[0pt][0pt]{$\bar{\Delta}$}  &  \raisebox{-0.3ex}[0pt][0pt]{ER} \\\hline
BZ-SCR & \textbf{0.2907} & \textbf{0.3409} & \textbf{0.1133} & \textbf{0.4664} \\\hline
Boosting & 0.3305   &   0.3704   &  0.1277  &  0.4966 \\\hline
SJE  &   0.3716   &   0.4239   &   0.1286   &   0.4988 \\\hline
ALE  &   0.3394   &   0.3875   &   0.1401   &   0.4954 \\\hline
NR   &   0.4191   &   0.5100   &   0.1661   &   0.6478 \\\hline
CCA  &   0.3945   &   0.4702   &   0.2614   &   0.6860 \\\hline
\end{tabular}
\end{table}

Table~\ref{tab_errrate} shows the $\bar{\Delta}$ (mean of $\Delta$) and the ER (error rate) performances of BZ-SCR and the comparative methods, where ``Boosting" is the boosting-only version of BZ-SCR (fixing all $s=\textbf{1}_N$). The best results are shown in bold face. Since SJE and ALE also involve a $\Delta$ loss in learning, we show their ER results as the better one between two forms of $\Delta$: the same as BZ-SCR (Eq.~(\ref{DeltaAWA})/(\ref{DeltaCUB})) and the uniform 0-1 error. The shown results of BZ-SCR are achieved with $\nu/N={10}^{-4},~\beta/N=0.4$ for \textit{AWA}, and $\nu/N\in\{0.025,0.05\},~\beta/N\in\{0.1,0.2\}$ for \textit{CUB200}. We see that BZ-SCR has a better overall performance than the comparative methods, since BZ-SCR jointly addresses the issues of model effectiveness and model adaptation for ZSL, based on a self-adaptive SCR-regularized boosting optimization.

\begin{figure*}[!htp]
\centering
\makebox[\textwidth][c]{
\includegraphics[scale=0.5]{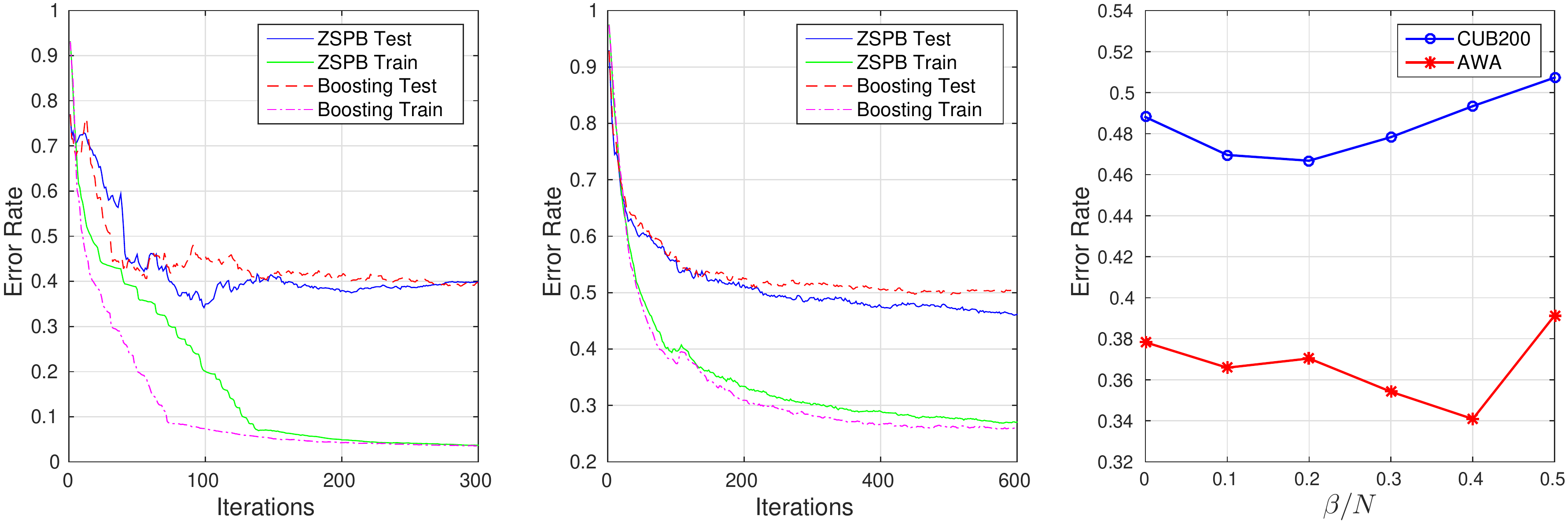}
}
\leftline{\small \hspace{2.4cm}(a) \textit{AWA}\hspace{4.875cm}(b) \textit{CUB200}\hspace{4.3cm}(c) Effect of SCR}
\caption{(a)(b) The error rates on the training and test set of BZ-SCR and boosting model w.r.t. the iterations, with $\nu/N={10}^{-4}, \beta/N=0.4$ for \textit{AWA}, and $\nu/N=0.025, \beta/N=0.2$ for \textit{CUB200}. BZ-SCR has a smaller gap between training and test curves. (c) The error rates of BZ-SCR w.r.t. $\beta/N$, with $\nu/N={10}^{-4}$ for \textit{AWA} and $\nu/N=0.025$ for \textit{CUB200}.}\label{fig_iter}
\end{figure*}

Furthermore, we show in Figure~\ref{fig_iter}(a)(b) the change of the error rates on the training and the test set w.r.t. the learning iterations of BZ-SCR and boosting model only, with the same parameter settings. We see that BZ-SCR generally has a smaller gap between the training curves and the test curves, which shows that BZ-SCR achieves a more stable learning process and a more adaptable model transfer. This is due to the smooth learning pace of BZ-SCR based on a self-controlled sample selection from easy reliable samples to hard confusing ones, instead of learning from the whole data batch as boosting does.

We investigate the efficacy of the semantic correlation regularization (SCR) for our model, and show the error rate results w.r.t. different $\beta$ values in Figure~~\ref{fig_iter}(c). From the figure we see that the integration of SCR is beneficial for the model performance. We empirically find that a better overall performance is generally achieved with $\beta/N$ around $0.1|\mathcal{Y}_S|/|\mathcal{Y}_T|$.

\section{Conclusion}
In this work, we study zero-shot learning (ZSL) as a transfer learning problem, and focus on its two key aspects, model effectiveness and model adaptation. We propose a unified framework, Boosted Zero-shot classification with Semantic Correlation Regularization (BZ-SCR), that simultaneously addresses these two aspects. We adopt the boosting strategy to learn an effective ensemble classifier, and devise a Semantic Correlation Regularization (SCR) to regularize the model with the inter-class semantic correlations for learning a semantically adaptable zero-shot model. Moreover, we embed a self-controlled sample selection into the framework to model the samples relations for a robust model generalization. By effectively balancing the SCR-regularized boosted model selection and the self-controlled sample selection, the BZ-SCR framework is capable of capturing both discriminative and adaptable feature-to-class semantic alignments, while ensuring the reliability and adaptability of the samples involved in learning. Thus, the proposed framework jointly considers the learning effectiveness, the cross-semantics adaptation, and the model robustness for learning a zero-shot classifier. The experiments on two classic ZSL image datasets verify the superiority of the proposed framework.


\section*{Acknowledgments}
This work was supported in part by the National Natural Science Foundation of China under Grants U1509206, 61472353, and 61672456, in part by the Alibaba-Zhejiang University Joint Institute of Frontier Technologies, and the Fundamental Research Funds for Central Universities in China.

\appendix

\section{Dual Derivation of Eqs.~(\ref{dual_opt}) and~(\ref{Q_rho})}\label{apdx_dual}
For Eq.~(\ref{opt_w}), we write its Lagrangian function:
\begin{align} \label{Lagran}
\mathcal{L}=&\sum_{i,r\in\mathcal{Y}_S}{s_iL(\rho_{ir})}+\sum_i{\beta s_iR(\rho_i^t)}+\nu\textbf{1}_K^Tw-\theta^Tw \\
&-\sum_{i,r}{Q_{ir}\left\{\rho_{ir}+F(x_i,y_i)-F(x_i,r)-\Delta(y_i,r) \right\}}, \notag
\end{align}
where $Q\in\mathbb{R}^{N\times C},~\theta\in\mathbb{R}^K~(\theta\geqslant0)$ are the Lagrangian multipliers for the equality constraints of $\rho_{ir}$ and the inequality constraint of $w\geqslant0$, respectively. Then we have:
\begin{align}
\frac{\partial\mathcal{L}}{\partial\rho_{ir}}=\left\{\begin{array}{rcl} s_iL'(\rho_{ir})-Q_{ir},~~~~~~~~~&r\in\mathcal{Y}_S\notag\\
\beta s_iL'({cov}_i)\frac{\partial{cov}_i}{\partial\rho_{ir}}-Q_{ir},~&r\in\mathcal{Y}_T
\end{array}\right. .
\end{align}

Since $L(\rho)=\ln{(1+e^\rho)},~{cov}_i=\text{cov}(\Delta_i^t,\rho_i^t)-D[\Delta_i^t]$, let ${\partial\mathcal{L}}/{\partial\rho_{ir}}=0$ (KKT conditions), we have:
\begin{align} \label{Q_rho2}
Q_{ir}=\left\{\begin{array}{rcl} \frac{s_i}{1+\exp{(-\rho_{ir})}},~~~~~~~~~~~~&r\in\mathcal{Y}_S\\
\frac{\Delta(y_i,r)-\text{E}[\Delta_i^t]}{|\mathcal{Y}_T|}\frac{\beta s_i}{1+\exp{(-{cov}_i)}},&r\in\mathcal{Y}_T
\end{array}\right. ,
\end{align}
which is exactly Eq.~(\ref{Q_rho}).

Furthermore, the derivative of $\mathcal{L}$ to $w$ is:
\begin{equation}
\frac{\partial\mathcal{L}}{\partial w}=\nu\textbf{1}_K-\theta-\sum_{ir}{Q_{ir}\left(h^{(iy_i)}-h^{(ir)}\right)}\notag,
\end{equation}
where $h^{(ir)}\in\mathbb{R}^K$ is the stacked vector of $h_j(x_i,r)$ for $j\in[1,K]$. Let ${\partial\mathcal{L}}/{\partial w}=0$, and based on $\theta\geqslant0$, we have:
\begin{equation} \label{dual_cons}
\sum_{ir}{Q_{ir}\left(h^{(iy_i)}-h^{(ir)}\right)}=\nu\textbf{1}_K-\theta\leqslant\nu\textbf{1}_K,
\end{equation}
which is the dual constraint of $Q$ for the dual problem.


By substituting $\rho_{ir}~(r\in\mathcal{Y}_S)$ from Eq.~(\ref{Q_rho2}) and $\theta$ from Eq.~(\ref{dual_cons}) into Eq.~(\ref{Lagran}), we have:
\begin{align}
\min_{w,\rho}{\mathcal{L}}=&\sum_{i,r\in\mathcal{Y}_S}{\left\{ s_i\ln{s_i}-H(Q_{ir})\right\}}+J(Q,\Delta) \notag\\
&+\sum_i{\min_{\rho_i^t}{\left\{\beta s_iR(\rho_i^t)-\langle Q_i^t,\rho_i^t\rangle\right\}}},    \notag
\end{align}
where $H(Q_{ir})=Q_{ir}\ln{Q_{ir}}+(s_i-Q_{ir})\ln{(s_i-Q_{ir})}$; $J(Q,\Delta)=\sum_{i,r}{Q_{ir}\Delta(y_i,r)}$; $Q_i^t,\rho_i^t\in\mathbb{R}^{|\mathcal{Y}_T|}$ are the stacked vectors of $Q_{ir}$ and $\rho_{ir}$ for $r\in\mathcal{Y}_T$, respectively. Based on the definition of Fenchel dual function, we have:
\begin{align}
&\min_{\rho_i^t}{\left\{\beta s_iR(\rho_i^t)-\langle Q_i^t,\rho_i^t\rangle\right\}}\notag\\
=&-\beta \max_{\rho_i^t}{\left\{\left\langle {Q_i^t}/{\beta},\rho_i^t\right\rangle-s_iR(\rho_i^t) \right\}}=-\beta R_{s_i}^\star\left({Q_i^t}/{\beta}\right), \notag
\end{align}
where $R_{s_i}^\star(\cdot)$ is the Fenchel dual function of $s_iR(\cdot)$.

Finally, based on the Lagrangian duality, the dual problem should be $\max\limits_Q{\min\limits_{w,\rho}\mathcal{L}}$, which is given by:
\begin{equation}
\max_Q{-\sum_{i,r\in\mathcal{Y}_S}{H(Q_{ir})}-\sum_i{\beta R_{s_i}^\star\left({Q_i^t}/{\beta}\right)}+J(Q,\Delta)}, \notag
\end{equation}
where we omit the constant term. Together with the constraint Eq.~(\ref{dual_cons}), the above optimization is equivalent to Eq.~(\ref{dual_opt}). Moreover, since $L(\cdot)$ is convex and ${cov}_i$ is linear to $\rho_i^t$, the primal objective function is convex and the duality gap is $0$.

\bibliographystyle{named}
\bibliography{ijcai17,mybibfile}

\end{document}